\pdfoutput=1
\documentclass[11pt]{article}
\usepackage[]{emnlp2021} 
\usepackage{times}
\usepackage{latexsym}
\usepackage[T1]{fontenc}
\usepackage[utf8]{inputenc}
\usepackage{microtype}

\usepackage{graphicx}
\graphicspath{ {./images/} }
\usepackage{multirow}
\usepackage{graphics}
\usepackage{parskip}
\usepackage{amssymb}
\usepackage{amsmath}
\usepackage{float}
\usepackage{enumitem}
\usepackage{booktabs}
\usepackage{xspace}
\usepackage{makecell}
\usepackage{blindtext}
\usepackage[noend,linesnumbered,ruled,vlined]{algorithm2e}

\newcommand{\our}{FineCL\xspace}

\title{Fine-grained Contrastive Learning for Relation Extraction}
\author{ William Hogan $\qquad$ Jiacheng Li $\qquad$ Jingbo Shang\thanks{$\,$ Corresponding author} \\
Department of Computer Science \& Engineering\\
University of California, San Diego \\
\texttt{\{whogan,j9li,jshang\}@ucsd.edu}
}

\begin{document}
\maketitle

\setlength{\abovedisplayskip}{6pt} 
\setlength{\belowdisplayskip}{6pt} 
\setlength{\belowcaptionskip}{0pt} 
\setlength{\algomargin}{10pt}

\begin{abstract}

Recent relation extraction (RE) works have shown encouraging improvements by conducting contrastive learning on silver labels generated by distant supervision before fine-tuning on gold labels. Existing methods typically assume all these silver labels are accurate and treat them equally; however, distant supervision is inevitably noisy---some silver labels are more reliable than others. In this paper, we propose fine-grained contrastive learning (\our) for RE, which leverages fine-grained information about which silver labels are and are not noisy to improve the quality of learned relationship representations for RE. We first assess the quality of silver labels via a simple and automatic approach we call ``learning order denoising,'' where we train a language model to learn these relations and record the order of learned training instances. We show that learning order largely corresponds to label accuracy---early-learned silver labels have, on average, more accurate labels than later-learned silver labels. Then, during pre-training, we increase the weights of accurate labels within a novel contrastive learning objective. Experiments on several RE benchmarks show that \our makes consistent and significant performance gains over state-of-the-art methods. 
\end{abstract}         

\begin{figure*}[t]
    \centering
    \includegraphics[width=1.0\textwidth]{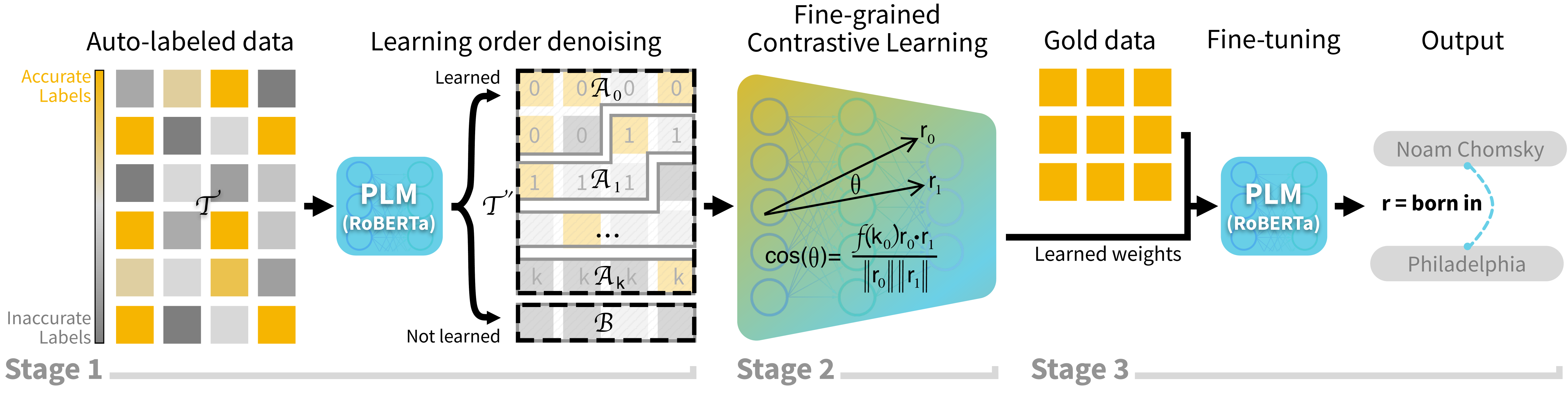}
    \caption{
    The \our framework has three stages: 
    Stage 1: we use distantly supervised data ($\mathcal{T}$) to train a PLM via cross-entropy to collect ordered subsets of learned ($\mathcal{A}$) and not learned ($\mathcal{B}$) instances over $k$ epochs. Stage 2: function $f(k)$ weighs relation instances ($r_0, r_1$) relative to their learning order in a contrastive learning pre-training objective that uses cosine similarity to align similar relations. Stage 3: we adapt the model to a discriminative task.}
    \label{fig:fgcl}
\end{figure*}

\section{Introduction}\label{sec:introduction}
Relation extraction (RE), a subtask of information extraction, 
is a foundational task in Natural Language Processing (NLP). The RE task is to determine a linking relationship between two distinct entities from text, producing fact triples in the form [\textit{head}, \textit{relation}, \textit{tail}].
For example, reading the Wikipedia page on Noam Chomsky, we learn that Noam was ``born to Jewish immigrants in Philadelphia,'' which corresponds to the fact triple [\textit{Noam Chomsky}, \textit{born in}, \textit{Philadelphia}]. Fact triples play a key role in downstream NLP tasks such as question-answering, search queries, dialog systems, and knowledge-graph completion~\cite{Xu2016QuestionAO, Lin2015LearningEA, Madotto2018Mem2SeqEI, Hogan2021AbstractifiedML, Li2014CoREAC}.

Current state-of-the-art RE models leverage a two-phase training: a self-supervised pre-training followed by a supervised fine-tuning. Popular pre-trained language models (PLM) such as BERT~\cite{Devlin2019BERTPO} and RoBERTa~\cite{Liu2019RoBERTaAR} feature a generic pre-training objective, namely masked language modeling (MLM), that allows them to generalize to various downstream tasks. However, recent RE works have shown impressive performance gains by using a pre-training objective designed specifically for relation extraction~\cite{BaldiniSoares2019MatchingTB, Peng2020LearningFC, Qin2021ERICAIE}. 

Recently, \citet{Peng2020LearningFC} and \citet{Qin2021ERICAIE} used a contrastive learning loss function to learn relationship representations during pre-training. However, RE-specific pre-training requires large amounts of automatically labeled data obtained trough distant supervision for RE~\cite{Mintz2009DistantSF} which is inherently noisy---not all labels from distantly supervised data are correct. 
\citet{Gao2021ManualEM} manually examined distantly supervised relation data and found that a significant ratio, 53\%, of the assigned labels were incorrect. Furthermore, distantly supervised labels can go beyond ``correct'' or ``incorrect''---they can have multiple levels of correctness. Consider the following sentences: 
\begin{enumerate}[nosep,leftmargin=*]
\small
\vspace{-1mm}
    \item ``\textit{Noam Chomsky} was born in \textit{Philadelphia}.''
    \item ``\textit{Noam Chomsky} gave a presentation in \textit{Philadelphia}.''
    \item ``Raised in the streets of \textit{Philadelphia}, \textit{Noam Chomsky}...''
\vspace{-1mm}
\end{enumerate}

Pairing this text with the Wikidata knowledge graph~\cite{wikidata}, distant supervision labels each sentence as a positive instance of [\textit{Noam Chomsky}, \textit{born in}, \textit{Philadelphia}]; however, only sentence (1) adequately expresses the relationship ``born in.'' Sentence (2) is incorrectly labeled, and sentence (3) is, arguably, semi-accurate since one may infer that someone was born in the same place they were raised. Conventional contrastive learning for RE does not account for differences in label accuracy---it treats all instances equally. This can be problematic when learning robust and high-quality relationship representations. 

This paper proposes a noise-aware contrastive pre-training, Fine-grained Contrastive Learning (\our) for RE, that leverages additional fine-grained information about which instances are and are not noisy to produce high-quality relationship representations. Figure~\ref{fig:fgcl} illustrates the end-to-end data flow for the proposed \our method. We first assess the noise level of all distantly supervised training instances and then incorporate such fine-grained information into the contrastive pre-training. Less noisy, or clean, training instances are weighted more relative to noisy training instances. We then fine-tune the model on gold-labeled data. As we demonstrate in this work, this approach produces high-quality relationship representations from noisy data and then optimizes performance using limited amounts of human-annotated data.

There are several choices of methods to assess noise levels. We select a simple yet effective method we call ``learning order denoising'' that does not require access to human annotated labels. We train an off-the-shelf language model to predict relationships from distantly supervised data and we record the order of relation instances learned during training. We show that the order in which instances are learned corresponds to the label accuracy of an instance: accurately labeled relation instances are learned first, followed by noisy, inaccurately labeled relation instances.

We leverage learning-order denoising to improve the relationship representations learned during pre-training by linearly projecting the weights of each relation instance corresponding to the order in which the instance was learned. We apply higher weights to relation instances learned earlier in training relative to those learned later in training. We use these weights to inform a contrastive learning loss function that learns to group instances of similar relationships.

We compare our method to leading RE pre-training methods and observe an increase in performance on various downstream RE tasks, illustrating that \our produces more informative relationship representations.

The contributions of this work are the following:
\begin{itemize}[nosep,leftmargin=*]
    \item We demonstrate that learning-order denoising is an effective and automatic method for denoising distantly labeled data.
    \item Applying a denoising strategy to a contrastive learning pre-training objective creates more informative representations, improving performance on downstream tasks.
    \item We openly provide all code, trained models, experimental settings, and datasets used to substantiate the claims made in this paper.\footnote{\url{https://github.com/wphogan/finecl}}
\end{itemize}

\section{Related Work}\label{sec:related_work}
Early RE methods featured pattern-based algorithms~\cite{Califf1997RelationalLO} followed by advanced statistical-based RE methods~\cite{Mintz2009DistantSF, Riedel2010ModelingRA, Quirk2017DistantSF}. Advances in deep learning led to neural-based RE methods~\cite{Zhang2015RelationCV, Peng2017CrossSentenceNR, Miwa2016EndtoEndRE}. The transformer~\cite{Vaswani2017AttentionIA} enabled the development of wildly successful large pre-trained language models~\cite{gpt1, Devlin2019BERTPO, Liu2019RoBERTaAR}. At the time of writing, all current leading models in RE\footnote{\url{https://paperswithcode.com/task/relation-extraction}} leverage large pre-trained language models via a two-step training methodology: a self-supervised pre-training followed by a supervised fine-tuning~\cite{Xu2021EntitySW, Xiao2021SAISSA}. 

%
%
\begin{table}[t]
\centering
\resizebox{\columnwidth}{!}{
\begin{tabular}{l | c c c c } 
    \hline
    \rule{0pt}{3ex}
    \rule{-3pt}{3ex}
        & Base Lang. Model  & Pre-train objective   & R\textsubscript{D}    & E\textsubscript{D}    \\ \hline 
    \rule{-3pt}{3ex}
    BERT                            & BERT      & MLM           & $\times$        &  $\times$ \\
    RoBERTa                         & RoBERTa   & MLM           & $\times$        &  $\times$ \\
    MTB                             & BERT      & DPS           & $\checkmark$	 &  $\times$ \\
    CP                              & BERT      & CL + MLM      & $\checkmark$	 &  $\times$ \\
    ERICA\textsubscript{BERT}       & BERT      & CL + MLM      & $\checkmark$	 &  $\checkmark$	 \\
    ERICA\textsubscript{RoBERTa}    & RoBERTa   & CL + MLM      & $\checkmark$	 &  $\checkmark$	 \\
    WCL                             & BERT      & WCL + MLM      & $\checkmark$	 &  $\times$ \\
    \our                            & RoBERTa   & \our + MLM    & $\checkmark$	 &  $\checkmark$	 \\ \hline
\end{tabular}} 
\caption{A comparison of RE pre-training methods highlighting the pre-training objective: Mask Language Modeling (MLM), Dot Product Similarity (DPS), Contrastive Learning (CL), Weighted Contrastive Learning (WCL), and Fine-grained Contrastive Learning (\our). R\textsubscript{D} denotes the presence of relation discrimination in the loss function, and E\textsubscript{D} denotes the presence of entity discrimination in the loss function.}
\label{tab:method_compare}
\end{table}

Building on BERT~\cite{Devlin2019BERTPO}, \citet{BaldiniSoares2019MatchingTB} proposed MTB, a model featuring a pre-training objective explicitly designed for the task of relation extraction. MTB uses dot product similarly to align pairs of randomly masked entities during pre-training. Its success inspired the development of subsequent RE-specific pre-training methods~\cite{Peng2020LearningFC, Qin2021ERICAIE}. \citet{Peng2020LearningFC} demonstrated the effectiveness of contrastive learning used to develop relationship representations during pre-training. Their model, named ``CP,'' featured a pre-training objective that combined a relation discrimination task with BERT's masked language modeling (MLM) task. Their work inspired ERICA~\cite{Qin2021ERICAIE}, which expanded the contrastive learning pre-training objective to include entity and relation discrimination, as well as MLM. 

\citet{wcl} is a recent extension of \citet{Peng2020LearningFC} that proposes a weighted contrastive learning (WCL) method for RE. The authors first fine-tune BERT to predict relationships using gold training data and then use the fine-tuned model to predict relationships from distantly labeled data. Next, they use the softmax probability of each prediction as a confidence value which they then apply to a weighted contrastive learning function used for pre-training. Lastly, they fine-tune the WCL model on gold training data.  

Our work is an extension of ERICA. We introduce a more nuanced RE contrastive learning objective that leverages additional, fine-grained data about which instances are high-quality training signals. Table~\ref{tab:method_compare} qualitatively compares recent pre-training methods used for RE. 

\section{Methods}\label{sec:methods}
FineCL for RE consists of three discrete stages: learning order denoising, contrastive pre-training, and supervised adaptation. 

\subsection{Learning Order Denoising}\label{sec:learning_order}
For learning order denoising, we automatically label large amounts of training data via distant supervision for RE~\cite{Mintz2009DistantSF} which we use to train a PLM to predict relation classes using multi-class cross-entropy loss. 
\begin{equation}\label{eq:cross_entropy}
\begin{aligned}
\mathcal{L}_{\mathrm{CE}}=-\sum_{i=1}^{N} y_{o,i} \cdot \log \left(p\left(y_{o,i}\right)\right)
\end{aligned}
\end{equation}
Where the number of classes $N$ is the number of relation classes plus one for \textit{no relation}, $y$ is a binary indicator that is $1$ if and only if $i$ is the correct classification for observation $o$, and $p(y_{o,i})$ is the Softmax probability that observation $o$ is of class $i$.

During training, we record the order of training instances learned. We consider an instance ``learned'' upon the initial correct prediction. Likewise, an instance is ``not learned'' if the model fails to predict it correctly during training. Training instances are evaluated by batch within each epoch, exposing the model to all training data points the same number of times. We refer to this method as \textit{batch-based} learning order.

Thus, the PLM effectively becomes a mapping function that maps all training instances ($\mathcal{T}$) into two subsets: learned ($\mathcal{A}$) and not learned instances ($\mathcal{B}$) such that $\mathcal{A} \bigcup \mathcal{B}  = \mathcal{T}^\prime$ and $ \mathcal{A} \bigcap \mathcal{B} = \emptyset$.

\begin{figure}[t]
    \centering
    \includegraphics[width=0.48\textwidth]{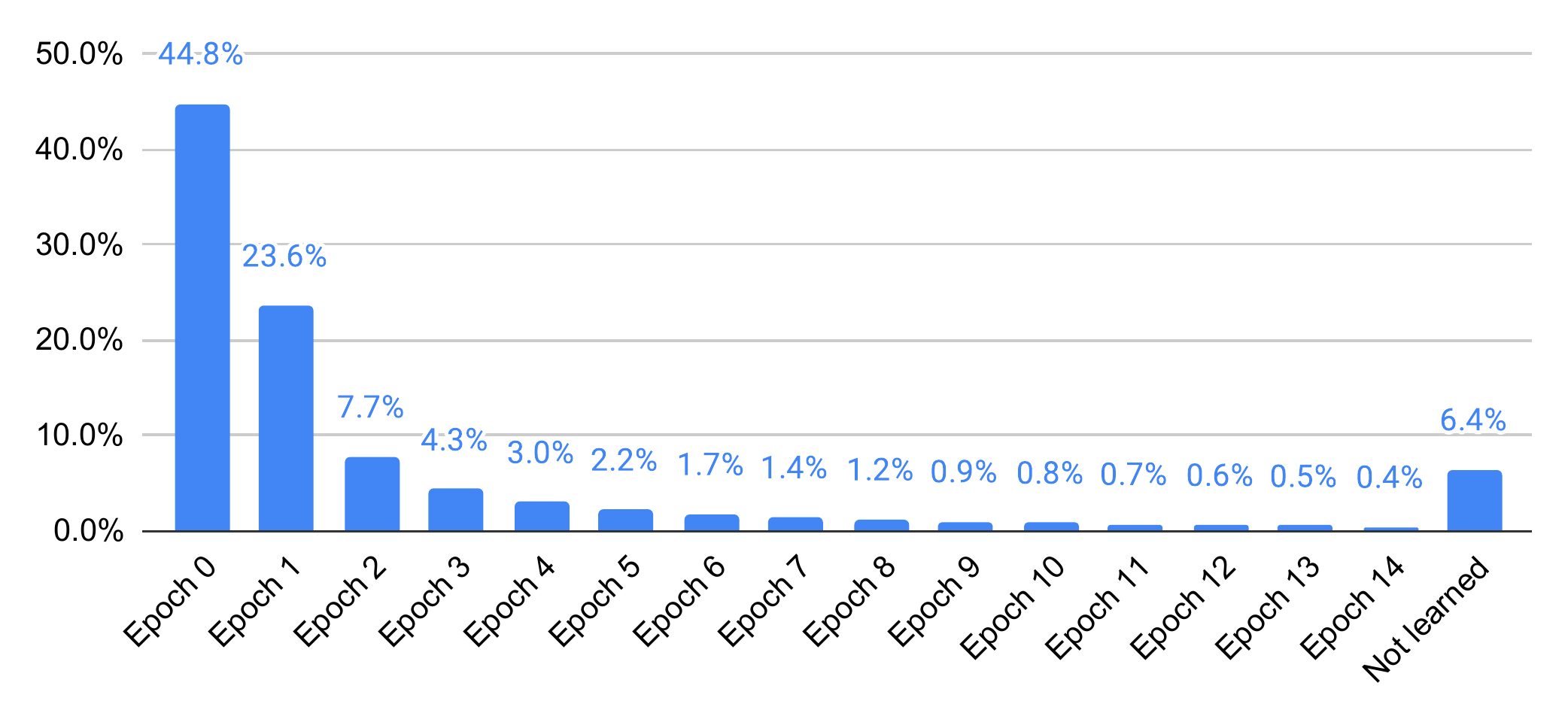}
    \caption{Percent of total training instances learned per epoch when recording \textit{batch-based} learning order on distantly labeled data from DocRED.}
    \label{fig:ratio_learned}
\end{figure}

The set of learned instances $\mathcal{A}$ is further divided into non-intersecting subsets of learned instances $\mathcal{A}_{1}$ through $\mathcal{A}_{k}$ where $k$ corresponds to the epoch in which an instance is learned.
\begin{gather}
\mathcal{A}_{0} \bigcup \mathcal{A}_{1} ... \bigcup \mathcal{A}_{k} = \mathcal{A}\\
\mathcal{A}_{i} \bigcap \mathcal{A}_{j}=\emptyset \text { for all } i \neq j
\end{gather}
We use $k=15$ epochs, resulting in $k+1$ subsets of instances---$k$ subsets of learned instances plus one subset of not learned instances. Figure~\ref{fig:ratio_learned} shows the percent of total training instances learned per epoch during this phase on the DocRED~\cite{docred} distantly labeled training set which contains 100k documents, 1.5M intra- and inter-sentence relation instances, and 96 relation types (not including no relation). 

More challenging relation classes may be underrepresented within the set of learned instances. Such minority classes can be problematic during pre-training since unlearned instances are weighted less than learned ones, presenting a challenge for the model to learn informative representations for minority classes. To account for this, we ensure that at least $P\%$ of instances of each relation class is contained within the set of learned instances. During training, we set $P=50$ and observed that 2\% of relation classes are underrepresented within the set of learned instances. We upsample underrepresented classes by randomly selecting unlearned instances from the corresponding class, placing them into one of the $k$ subsets of learned instances $\mathcal{A}$. See Figure~\ref{fig:complete_learning_order} in the Appendix for a detailed chart showing the ratio of learned instances by relation class in each epoch. 

Learning order metadata is then inserted into the original training data $\mathcal{T}$, creating a modified training set $\mathcal{T}^\prime$ used for the contrastive pre-training.

\subsection{Contrastive Pre-training}\label{sec:experiments}

This section introduces our pre-training method to learn high-quality entity and relation representations. We first construct informative representation for entities and relationships which we use to implement a three-part pre-training objective that features entity discrimination, relation discrimination, and masked language modeling. 

\subsubsection{Entity \& Relation Representation}\label{sec:entity_relation_repr}
We construct entity and relationship representations following ERICA~\cite{Qin2021ERICAIE}. For the document $d_i$, we use a pre-trained language model to encode $d_i$ and obtain the hidden states $\{\mathbf{h}_1, \mathbf{h}_2,\dots,\mathbf{h}_{|d_i|}\}$. Then, \textit{mean pooling} is applied to the consecutive tokens in entity $e_j$ to obtain entity representations. Assuming $n_{\mathrm{start}}$ and $n_{\mathrm{end}}$ are the start index and end index of entity $e_j$ in document $d_i$, the entity representation of $e_j$ is represented as:
\begin{equation}
    \mathbf{m}_{e_j} = \mathrm{MeanPool}(\mathbf{h}_{n_{\mathrm{start}}},\dots,\mathbf{h}_{n_{\mathrm{end}}}),
\end{equation}
To form a relation representation, we concatenate the representations of two entities $e_{j1}$ and $e_{j2}$: $\mathbf{r}_{j1j2} = [\mathbf{e}_{j1};\mathbf{e}_{j2}]$. 

\subsubsection{Entity Discrimination}
For entity discrimination, we use the same method described in ERICA. The goal of entity discrimination ($\mathrm{E_D}$) is inferring the tail entity in a document given a head entity and a relation~\cite{Qin2021ERICAIE}. The model distinguishes the ground-truth tail entity from other entities in the text. Given a sampled instance tuple $t^i_jk = (d_i, e_{ij}, r^i_{jk}, e_{ik})$, our model is trained to distinguish the tail entity $e_{ik}$ from other entities in the document $d_i$. Specifically, we concatenate the relation name of $r^i_{jk}$, the head entity $e_{ij}$ and a special token \texttt{[SEP]} in front of $d_i$ to get $d^*_i$. Then, we encode $d^*_i$ to get the entity representations using the method from Section~\ref{sec:entity_relation_repr}. The contrastive learning objective for entity discrimination is formulated as:

\resizebox{.48 \textwidth}{!} 
{
    $ \mathcal{L}_{\mathrm{E_D}}=-\sum_{t_{j k}^{i} \in \mathcal{T}^\prime} \log \frac{\exp \left(\cos \left(\mathbf{e}_{i j}, \mathbf{e}_{i k}\right) / \tau\right)}{\sum_{l=1, l \neq j}^{\left|\mathcal{E}_{i}\right|} \exp \left(\cos \left(\mathbf{e}_{i j}, \mathbf{e}_{i l}\right) / \tau\right)}$
}

where $\mathrm{cos}(\cdot,\cdot)$ denotes the cosine similarity between two entity representations and $\tau$ is a temperature hyper-parameter.

\subsubsection{Relation Discrimination}
To effectively learn representation for downstream task relation extraction, we conduct a Relation Discrimination ($\mathrm{R_D}$) task during pre-training. $\mathrm{R_D}$ aims to distinguish whether two relations are semantically similar~\cite{Qin2021ERICAIE}. Existing methods~\cite{Peng2020LearningFC, Qin2021ERICAIE} require large amounts of automatically labeled data from distant supervision which is noisy because not all sentences will adequately express a relationship.

In this case, the learning order can be introduced to make the model aware of the noise level of relation instances. To efficiently incorporate learning order into the training process, we propose fine-grained, noise-aware relation discrimination.

In this new method, the noise level of all distantly supervised training instances controls the optimization process by re-weighting the contrastive objective. Intuitively, the model should learn more from high-quality, accurately labeled training instances than noisy, inaccurately labeled instances. Hence, we assign higher weights to earlier learned instances from the learning order denoising stage.

In practice, we sample a tuple pair of relation instance $t_A = (d_A, e_{A_1}, r_A, e_{A_2}, k_A)$ and $t_B = (d_B, e_{B_1}, r_B, e_{B_2}, k_B)$ from $\mathcal{T}^\prime$ and $r_A = r_B$, where $d$ is a document; $e$ is a entity in $d$; $r$ is the relationship between two entities and $k$ is the first learned order introduced in Section~\ref{sec:learning_order}. Using the method mentioned in Section~\ref{sec:entity_relation_repr}, we obtain the positive relation representations $\mathbf{r}_{t_A}$ and $\mathbf{r}_{t_B}$. To discriminate positive examples from negative ones, the fine-grained $\mathrm{R_D}$ is defined as follows:
\begin{equation}
\begin{aligned}
\mathcal{L}_{\mathrm{R_D}} &=-\sum_{t_{A}, t_{B} \in \mathcal{T}^\prime}f(k_A)\log \frac{\exp \left(\cos \left(\mathbf{r}_{t_{A}}, \mathbf{r}_{t_{B}}\right) / \tau\right)}{\mathcal{Z}}, \\
\mathcal{Z} &=\sum_{t_{C} \in \mathcal{T}^\prime /\left\{t_{A}\right\}}^{N} f(k_C)\exp \left(\cos \left(\mathbf{r}_{t_{A}}, \mathbf{r}_{t_{C}}\right) / \tau\right) \\
\end{aligned}
\end{equation}
where $\mathrm{cos}(\cdot,\cdot)$ denotes the cosine similarity; $\tau$ is the temperature; $N$ is a hyper-parameter and $t_{C}$ is a negative instance ($r_A \neq r_C$) sampled from $\mathcal{T}^\prime$. Relation instances $t_A$ and $t_C$ are re-weighted by function $f$ which is defined as:
\begin{equation}
    f(k) = \alpha^{\frac{k_{\mathrm{max}}-k}{k_{\mathrm{max}}- k_{\mathrm{min}}}},
\end{equation}
where $\alpha$ ($\alpha > 1$) is a hyper-parameter of the function $f$; $\mathrm{max}$ and $\mathrm{min}$ are maximum and minimum first-learned order, respectively. We increase the weight of negative $t_C$ if it is a high-quality training instance (i.e.,~$k$ is small). Because all positives and negatives are discriminated from instance $t_A$, we control the overall weight by the learning order $k_A$.


\subsubsection{Overall Objective}
We include the MLM task~\cite{Devlin2019BERTPO} to avoid catastrophic forgetting of language understanding~\cite{McCloskey1989CatastrophicII} and construct the following overall objective for \our:
\begin{equation}
\mathcal{L}_{FineCL} = \mathcal{L}_{E_D} + \mathcal{L}_{R_D} + \mathcal{L}_{MLM} 
\end{equation}

\subsection{Supervised Adaptation}\label{subsec:supervised_adaptation}
The primary focus of our work is to improve relationship representations learned during pre-training and, in doing so, improve performance on downstream RE tasks. To illustrate the effectiveness of our pre-training method, we use cross-entropy loss, as described in equation~\ref{eq:cross_entropy}, to fine-tune our pre-trained \our model on document-level and sentence-level RE tasks.          
\section{Experiments}
%
%
\subsection{Learning Order as Noise Level Hypothesis}\label{sec:learning_order_hypothesis}
We first seek to confirm our hypothesis that the learning order automatically orders distantly supervised data from clean, high-quality instances to noisy, low-quality instances. However, given the large amount of pre-training data, statistically significant confirmation via manual annotation is prohibitively expensive. So, we devise the following experiment to test our hypothesis in lieu of a significant manual annotation effort.

We begin with the assumption that a model trained on a dataset without noise will perform better than a model trained on a dataset with noise. Suppose learning order denoising successfully orders instances relative to their noise; then, we should observe a boost in performance by training on a subset of early-learned instances compared to a model trained on the complete, noisy dataset.

As reported by~\citet{Gao2021ManualEM}, up to 53\% of relation instances labeled via distant supervision are incorrect. Using this estimation, we attempt to use learning order denoising to remove the roughly 50\% of instances that are noisy instances from the DocRED's distantly supervised training set. To do this, we first obtain the learning order of relation instances using the methodology described in Section~\ref{sec:learning_order}. Without loss of generalization, we choose RoBERTa~\cite{Liu2019RoBERTaAR}, specifically the \textit{roberta-base} checkpoint\footnote{\url{https://huggingface.co/roberta-base}}, as the base model to develop the order of learned instances.

\begin{table}[t]
\centering
\resizebox{\columnwidth}{!}{
\begin{tabular}{l c c c } 
\toprule
    \textbf{Learning order}  & \textbf{Training set} & \textbf{Training set size} & \textbf{F1} \\ \midrule
    \rule{-3pt}{3ex}
        None                    & $\mathcal{T}$                             & 100\%      & 45.8 \\
    \rule{-3pt}{3ex}
        Batch-based             & $\mathcal{T}_{\mathcal{A}_0^B}$  & 45.0\%     & \textbf{46.6}\\
    \rule{-3pt}{3ex}
        Epoch-based             & $\mathcal{T}_{\mathcal{A}_0^E}$  & 64.9\%     & 46.0 \\
\bottomrule
\end{tabular}} 
\caption{Results comparing performance on the DocRED test set using trimmed sets of distantly supervised training data. The \textit{batch-based} and \textit{epoch-based} training sets consist of training instances determined by the instances learned within the first epoch using the respective learning order collection methods.}
\label{tab:batch_vs_epoch_based}
\end{table}

We observe that the set of training instances learned via \textit{batch-based} learning order in the first epoch, $\mathcal{A}_0^B$, consists of 45\% of the total training instances. We use $\mathcal{A}_0^B$ to construct a trimmed training set $\mathcal{T}_{\mathcal{A}_0^B}$. We then compare performance in two settings: (1) RoBERTa trained with the complete distantly supervised training dataset $\mathcal{T}$ and (2) RoBERTa trained on the trimmed, denoised training data $\mathcal{T}_{\mathcal{A}_0^B}$. Table~\ref{tab:batch_vs_epoch_based} reports the results of this experiment. Significantly, the denoised training set consisting of only 45\% training data outperforms the baseline model. 

We also conduct an informal manual analysis of the learning order. We randomly selected 120 instances from the first six training epochs---60 correctly, and 60 incorrectly predicted instances. We find that 93\% of the correct predictions have accurate labels within the first three epochs. However, in epochs 4 through 6, label accuracy drops to 53\% among correct predictions. Furthermore, we find a relatively low label accuracy of 50\% from the first three epochs of incorrect predictions, illustrating that the model struggles to learn noisy instances compared to clean instances early in training. We use these results and the results presented in Table~\ref{tab:batch_vs_epoch_based} to argue that learning order successfully orders instances from clean, high-quality to noisy, low-quality instances.

%
%
\subsection{Learning Order: Batch- vs. Epoch-based}
We experiment with two methods of collecting learning order data: \textit{batch-based} and \textit{epoch-based} (see Appendix~\ref{app:bb_vs_eb} for pseudo-code describing these methods). \\
\textbf{Batch-based:} As previously mentioned, for \textit{batch-based} learning order we collect learned instances per batch across each epoch during training. However, we recognize that this may bias the set of learned instances by the random batch for which they are selected. For example, accurately labeled relation instances selected for the first few batches during training may not be predicted correctly because the model has not learned much. \\
\textbf{Epoch-based:} To reduce potential selection order bias from \textit{batch-based} learning order, we experiment with \textit{epoch-based} learning order 
by evaluating the model on the entire training set at the end of each epoch. We rerun the experiment detailed in Section~\ref{sec:learning_order_hypothesis} using \textit{epoch-based} learning order to construct the trimmed dataset $\mathcal{T}_{\mathcal{A}_0^E}$ and present the results in Table~\ref{tab:batch_vs_epoch_based}.

Using \textit{epoch-based} learning order, we observe that the model learns 64.9\% of the training instances within the first epoch, an increase compared to the 45.0\% of learned instances from \textit{batch-based} learning order. However, training RoBERTa on the \textit{epoch-based} training subset, we obtain an F1 score of 46.0, which under-performs relative to the 46.6 F1 score from the \textit{batch-based} learning order experiment. We hypothesize that, while \textit{epoch-based} learning order may capture more learned instances, it leads to noisier instances leaking into the sets of learned data because the model is more prone to simply memorizing noisy labels encountered previously in the epoch. 

Note that we do not use DocRED's human-annotated training data in these learning order experiments. Instead, we train on the distantly supervised training data and test on human-annotated data. This is done to assess the quality of the various subsets of distantly labeled data. It is why the performance of these tests is considerably lower than the results from the experiments in Section~\ref{subsec:experiments_relation_extraction} that leverage human-annotated training data.

%
%
\subsection{Pre-training Details}
To ensure a fair comparison and highlight the effectiveness of \our, we align our pre-training data and settings to those used by ERICA. The ERICA pre-training dataset is constructed using distant supervision for RE by pairing documents from Wikipedia (English) with the Wikidata knowledge graph. This distantly labeled dataset creation method mirrors the method used to create the distantly labeled training set in DocRED but differs in that it is much larger and more diverse. It contains 1M documents, 7.2M relation instances, and 1040 relation types compared to DocRED's 100k documents, 1.5M relation instances, and 96 relation types (not including \textit{no relation}). Additional checks are performed to ensure no fact triples overlap between the training data and the test sets of the various downstream RE tasks. Detailed pre-training settings can found in Appendix~\ref{app:pretrain}.

%
%
\begin{table}[t]
\centering
\resizebox{\columnwidth}{!}{
\begin{tabular}{l | c c | c c | c c } 
    \hline
    \rule{0pt}{3ex}
    \textbf{Size}               & \multicolumn{2}{c}{1\%}   & \multicolumn{2}{c}{10\%}  & \multicolumn{2}{c}{100\%} \\  \hline
    \rule{0pt}{3ex}
    \textbf{Metrics}            & F1     & IgF1      & F1    & IgF1      & F1    & IgF1  \\ \hline 
    \rule{-2pt}{3ex}
    CNN*                         & -      & -         & -     & -         & 42.3  & 40.3  \\ 
    BiLSTM*                      & -      & -         & -     & -         & 51.1  & 50.3  \\  \hline   
    \rule{-3pt}{3ex}
    HINBERT*                     & - & - & - & - & 55.6 & 53.7 \\
    CorefBERT*                   & \underline{32.8} & \textbf{31.2} & 46.0 & 43.7 & 57.0 & 54.5 \\
    SpanBERT*                    & 32.2 & 30.4 & 46.4 & 44.5 & 57.3 & 55.0 \\
    ERNIE*                       & 26.7 & 25.5 & 46.7 & 44.2 & 56.6 & 54.2 \\
    MTB*                         & 29.0 & 27.6 & 46.1 & 44.1 & 56.9 & 54.3 \\
    CP*                          & 30.3 & 28.7 & 44.8 & 42.6 & 55.2 & 52.7 \\
    BERT                        & 19.9 & 18.8 & 45.2 & 43.1 & 56.6 & 54.4 \\
    RoBERTa                     & 29.6 & 27.9 & 47.6 & 45.7 & 58.2 & 55.9 \\
    ERICA\textsubscript{BERT}   & 22.9 & 21.7 & 48.5 & 46.4 & 57.4 & 55.2 \\
    ERICA\textsubscript{RoBERTa}& 30.0 & 28.2 & \underline{50.1} & \underline{48.1} & \underline{59.1} & \underline{56.9} \\
    WCL\textsubscript{RoBERTa}  & 22.3 & 20.8 & 49.4 & 47.5 & 58.5 & 56.2 \\ \hline 
    \rule{-3pt}{3ex}
    \our                        & \textbf{33.2} & \textbf{31.2} & \textbf{50.3} & \textbf{48.3} & \textbf{59.5} & \textbf{57.1} \\ \hline
\end{tabular}} 
\caption{F1-micro scores reported on the DocRED test set. IgF1 ignores performance on fact triples in the test set overlapping with triples in the train/dev sets. (* denotes performance as reported in~\cite{Qin2021ERICAIE}; all other numbers are from our implementations).}
\label{tab:docred}
\end{table}

\subsection{Relation Extraction}\label{subsec:experiments_relation_extraction}

\textbf{Document-level RE}: To assess our framework's ability to extract document-level relations, we report performance on DocRED~\cite{docred}. We compare our model to the following baselines: (1) CNN~\cite{zeng-etal-2014-relation}, (2) BiLSTM \cite{lstm}, (3) BERT~\cite{Devlin2019BERTPO}, (4) RoBERTa~\cite{Liu2019RoBERTaAR}, (5) MTB~\cite{BaldiniSoares2019MatchingTB}, (6) CP~\cite{Peng2020LearningFC}, (7 \& 8) ERICA\textsubscript{BERT} \& ERICA\textsubscript{RoBERTa}~\cite{Qin2021ERICAIE}, (9) WCL~\cite{wcl}.  We fine-tune the pre-trained models on DocRED's human-annotated train/dev/test splits (see Appendix~\ref{app:docred} for detailed experimental settings). We implement WCL with identical settings from our other pre-training experiments and, for fair comparison, we use RoBERTa instead of BERT as the base model for WCL, given the superior performance we observe from RoBERTa in all other experiments. Table~\ref{tab:docred} reports performance across multiple data reduction settings (1\%, 10\%, and 100\%), using an overall F1-micro score and an F1-micro score computed by ignoring fact triples in the test set that overlap with fact triples in the training and development splits. We observe that \our outperforms all baselines in all experimental settings, offering evidence that \our produces better relationship representations from noisy data.

Given that learning-order denoising weighs earlier learned instances over later learned instances, \our may be biased towards easier, or common relation classes. The increase in F1-micro performance may result from improved predictions on common relation classes at the expense of predictions on rare classes. To better understand the performance gains, we also report F1-macro and F1-macro weighted in Table~\ref{tab:macro_docred}. The results show that \our outperforms the top baselines in both F1-macro metrics indicating that, on average, our method improves performance across all relation classes. However, the low F1-macro scores from all the models highlight an area for improvement---future pre-trained RE models should focus on improving performance on long-tail relation classes.

%
%
\begin{table}[t]
\centering
\resizebox{\columnwidth}{!}{
\begin{tabular}{l | c | c } 
    \hline
    \rule{0pt}{3ex}
    \textbf{Metric}                 & F1-macro     & F1-macro-weighted \\ \hline 
    \rule{-3pt}{3ex}
    BERT                            & 37.3              & 54.9 \\
    RoBERTa                         & 39.6              & 56.9 \\
    ERICA\textsubscript{BERT}       & 37.9              & 55.8 \\
    ERICA\textsubscript{RoBERTa}    & \underline{40.1 } & \underline{57.8} \\ 
    WCL\textsubscript{RoBERTA}      & 39.9              & 57.2 \\ \hline
    \rule{-3pt}{3ex}
    \our                            & \textbf{40.7 }     & \textbf{58.2} \\ \hline
\end{tabular}} 
\caption{F1-macro and F1-macro-weighted scores reported from the DocRED test set.}
\label{tab:macro_docred}
\end{table}

%
%
\begin{table}[t]
\centering
\resizebox{\columnwidth}{!}{
\begin{tabular}{l | c c c | c c c } 
    \hline
    \rule{0pt}{3ex}
    \textbf{Dataset}               & \multicolumn{3}{c}{TACRED}   & \multicolumn{3}{c}{SemEval} \\  \hline
    \rule{-3pt}{3ex}
    \textbf{Size}               & 1\%     & 10\%      & 100\%    & 1\%      & 10\%    & 100\%  \\ \hline 
    \rule{-3pt}{3ex}
    MTB*                            & 35.7 & 58.8 & 68.2 & 44.2 & 79.2 & 88.2 \\
    CP*                             & 37.1 & 60.6 & 68.1 & 40.3 & 80.0 & \underline{88.5} \\ 
    BERT                            & 22.2 & 53.5 & 63.7 & 41.0 & 76.5 & 87.8 \\
    RoBERTa                         & 27.3 & 61.1 & 69.3 & 43.6 & 77.7 & 87.5 \\
    ERICA\textsubscript{BERT}       & 34.9 & 56.0 & 64.9 & 46.4 & 79.8 & 88.1 \\
    ERICA\textsubscript{RoBERTa}    & \underline{41.1} & \underline{61.7} & 69.5 & \underline{50.3} & \underline{80.9} & 88.4 \\ 
    WCL\textsubscript{RoBERTA}      & 37.6 & 61.3 & \underline{69.7} & 47.0 & 80.0 & 88.3 \\ \hline
    \rule{-3pt}{3ex}
    \our                & \textbf{43.7} & \textbf{62.7} & \textbf{70.3} & \textbf{51.2} & \textbf{81.0} & \textbf{88.7}\\ \hline
\end{tabular}} 
\caption{F1-micro scores reported from the TACRED and SemEval test sets (* denotes performance as reported in \cite{Qin2021ERICAIE}; all other numbers are from our implementations).}
\label{tab:tacred_semeval}
\end{table}

\textbf{Sentence-level RE}: To assess our framework's ability to extract sentence-level relations, we report performance on TACRED \cite{zhang2017tacred} and SemEval-2010 Task 8 \cite{hendrickx-etal-2010-semeval}. We compare our model to MTB, CP, BERT, RoBERTa, ERICA\textsubscript{BERT}, ERICA\textsubscript{RoBERTa}, and WCL (see Appendix \ref{app:semeval_tacred} for detailed experimental settings). Table \ref{tab:tacred_semeval} reports F1 scores across multiple data reduction settings (1\%, 10\%, 100\%). Again, we observe that \our outperforms all baselines in all settings.      
\section{Ablation Studies}\label{sec:analysis}
We conduct a suite of ablation experiments to understand how learning order denoising affects the quality of relationship representations learned during pre-training. We note that the \our method is identical to ERICA when we remove fine-grained data and treat all instances equally. As such, ERICA can be considered an ablation experiment of \our without fine-grained data.

\subsection{Learning Order Epochs}
In our first ablation experiment, we vary the number of training epochs ($k$) used to obtain learning order data to determine how the different amounts of \textit{batch-based} learning order data affect pre-training. We test $k=\{1, 3, 5, 10, 15\}$ as well as a baseline that does not use learning order denoising. To reduce the high computational requirements for pre-training, we use a shortened pre-training for these experiments where we pre-train for 1000 training steps compared to the full 6000 step training used for our main experiments. We then fine-tune the models using the same settings described in Section~\ref{subsec:experiments_relation_extraction}. Notably, our pre-trained model trained at 1000 steps achieves an F1 score of 59.0, which is reasonably close to the 59.5 F1 score from the \our trained for 6000 steps. Table~\ref{tab:docred_ablation} contains the results from this ablation experiment. We observe that $k=15$ epochs of learned instances produce the best performance, indicating that a more extensive set of learned instances produces better relationship representations.

\begin{table}[t]
\centering
\resizebox{\columnwidth}{!}{
\begin{tabular}{l c c c } 
    \hline
    \textbf{Epochs of learning order data} & \textbf{\% Learned} & \textbf{F1} & \textbf{IgF1} \\ \hline
    \rule{-3pt}{3ex}
    Baseline            & N/A   & 58.7 & 56.5 \\
    1 Epoch             & 45    & 58.6 & 56.4 \\
    3 Epochs            & 76    & 58.6 & 56.3 \\
    5 Epochs            & 83    & 58.7 & 56.5 \\
    10 Epochs           & 92    & 58.8 & 56.6 \\
    15 Epochs           & 94    & \textbf{59.0} & \textbf{56.7} \\ \hline
\end{tabular}
} 
\caption{Ablation experiment results on the DocRED test set with pre-trained models that use learning order data obtained with various training durations. Percent learned refers to the percent of training instances learned in the set of learned instances ($\mathcal{A}$). ``Baseline'' is a pre-trained model that does not leverage learning order (i.e., all instances are weighted equally during pre-training).}
\label{tab:docred_ablation}
\end{table}

\subsection{Different Learning Order Models}
\begin{figure}[t]
    \centering
    \includegraphics[width=0.48\textwidth]{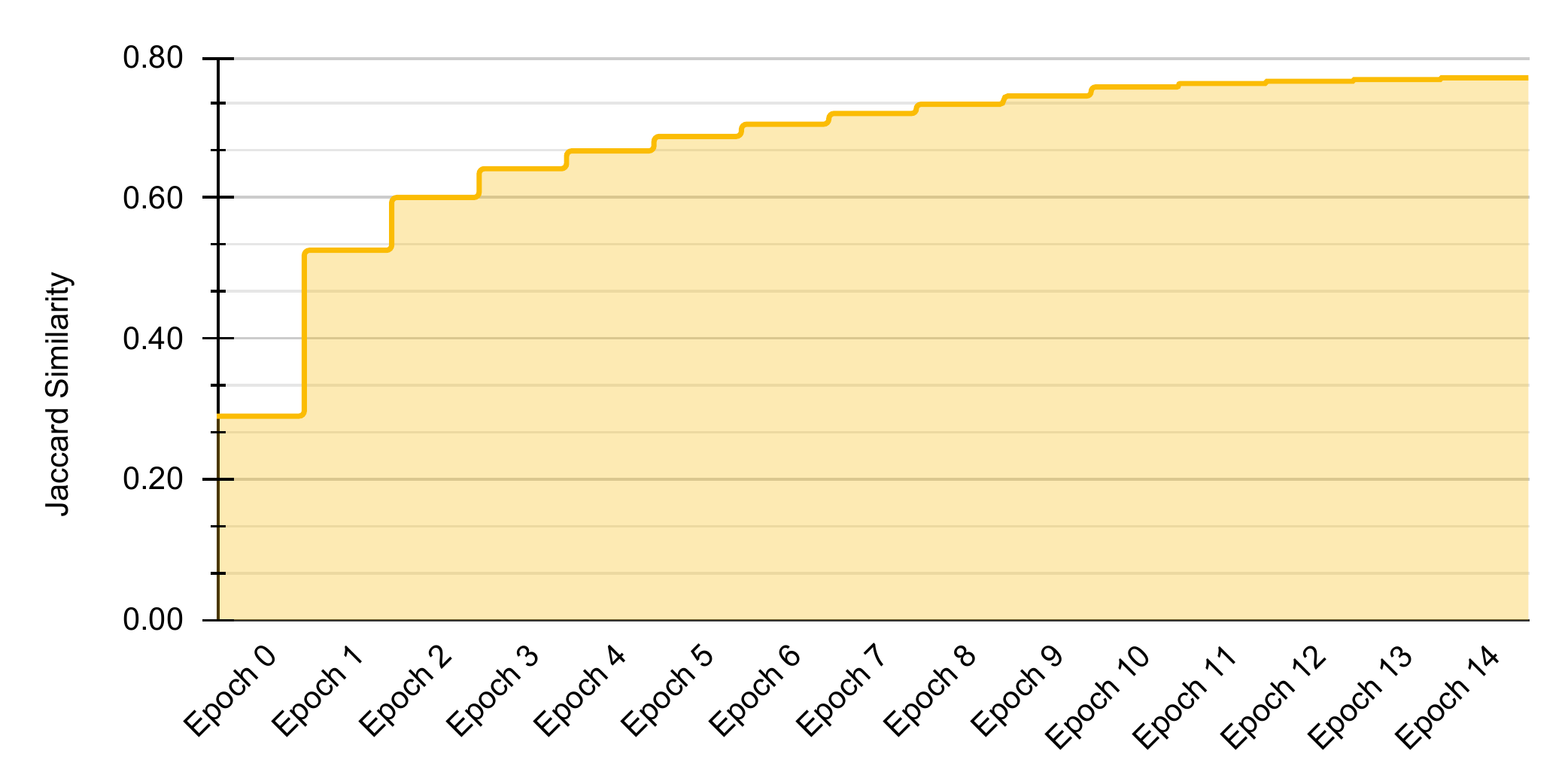}
    \caption{Cumulative Jaccard Similarity between sets of learned instances by epoch from RoBERTa and SSAN using distantly labeled training data from DocRED.}
    \label{fig:jaccard_sim}
\end{figure}

We chose the RoBERTa base model for the first stage of our \our framework to reduce the adoption barrier for our methodology. Popular pre-trained models such as \textit{roberta-base} are easy to implement and require fewer resources compared to larger state-of-the-art (SOTA) RE models. However, given that RoBERTa is not a leading RE model, we seek to answer the question---how do sets of learned training instances differ between RoBERTa and the SOTA RE model?  At the time of writing, the leading RE model on DocRED\footnote{\url{https://paperswithcode.com/sota/relation-extraction-on-docred}} is the SSAN model~\cite{Xu2021EntitySW}. Therefore, we compare sets of learned instances from SSAN ($\mathcal{A}^S$) and RoBERTa ($\mathcal{A}^R$) by epoch ($k$) using a cumulative Jaccard Similarity Index:
\[ J(\mathcal{A}^R, \mathcal{A}^S)= \sum_{i=0}^{k} \frac{|\mathcal{A}_{i}^{R} \cap \mathcal{A}_{i}^S|}{|\mathcal{A}_{i}^R \cup \mathcal{A}_{i}^S|} \]

Figure~\ref{fig:jaccard_sim} plots the cumulative Jaccard Similarity Index (JSI) between sets of learned instances from RoBERTa and SSAN. The total cumulative JSI between the two models after $k=15$ epochs is 0.771, showing high similarity between sets of learned instances. While the sets are not perfectly aligned, we argue that this high similarity justifies using the smaller and more convenient RoBERTa model in determining learning order. We leave a more thorough examination of the differences in sets of learned instances obtained using various RE models to future work and present our findings as a proof of concept, demonstrating that obtaining learning order from relatively small and convenient language models is sufficient in improving representations learned during pre-training. 

\subsection{Performance Relative to Class Difficulty}

%
%
\begin{table}[t]
\centering
\begin{tabular}{l | c } 
    \hline
    \rule{0pt}{3ex}
    \textbf{Metric}                 & F1-micro \\ \hline 
    \rule{-3pt}{3ex}
    BERT                            & 32.9      \\
    RoBERTa                         & \underline{35.8}      \\
    ERICA\textsubscript{BERT}       & 34.7      \\
    ERICA\textsubscript{RoBERTa}    & 34.4 \\ 
    WCL\textsubscript{RoBERTA}      & 35.7      \\ \hline
    \rule{-3pt}{3ex}
    \our                            & \textbf{36.1}     \\ \hline
\end{tabular}
\caption{F1-micro scores on a subset of difficult relation classes from the DocRED dataset. }
\label{tab:difficult_classes}
\end{table}

As mentioned in Section~\ref{sec:learning_order}, it is possible that learning-order denoising biases the model to easier, common relation classes, as easier classes may be over-represented in the set of learned instances. To understand the effectiveness of our approach relative to class difficulty, we assess the end-to-end performance of \our on a set of difficult relation classes. 

We recognize that there are multiple ways to define a ``difficult'' relation class. Difficult classes can be classes with few training instances, classes with a significant number of inaccurate or semi-accurate labels, or classes that suffer from low overall accuracy after training completes. For this ablation study, we define the set of difficult relation classes as classes that attain relatively low accuracy from the training in Stage 1 of \our. We claim that any class which achieves less than 80\% accuracy after Stage 1 training completes is a ``difficult'' relation class. This subset of the lowest-performing classes from the DocRED dataset makes up 24\% of all the classes in the dataset. 

We compare the end-to-end performance of \our to baselines that do not leverage fine-grained contrastive learning on the set of difficult relation classes. Table~\ref{tab:difficult_classes} contains the results from this experiment. We observe that \our achieves an F1 score of 36.1\% on the subset of difficult classes compared to the best-performing baseline which achieves 35.8\%. We argue that these results, as well as the results from Table~\ref{tab:macro_docred}, offer evidence that the \our approach is capable of improving performance on both difficult classes as well as easy classes. However, the low overall performance from all models on difficult classes highlights an area for future work. 

\section{Conclusion}
In this work, we expand on contrastive learning for relation extraction by introducing Fine-grained Contrastive Learning for RE---a method that uses additional, fine-grained information about distantly supervised training data to improve relationship representations learned during pre-training. These improved representations lead to increases in performance across a variety of downstream RE tasks. 
This report shows that learning order denoising effectively and automatically orders distantly supervised training data from clean to noisy instances. In future work, we hope to explore the usefulness of this method when applied to manually annotated data where learning order may instead reflect the level of difficulty of training instances. This could be an easy and automatic way to introduce curricula learning within the fine-tuning training phase. We also intend to explore the pairing of other denoising methods with \our.

\section*{Acknowledgements}
Thank you to the anonymous reviewers for their thoughtful feedback. Our work is sponsored in part by National Science Foundation Convergence Accelerator under award OIA-2040727 as well as generous gifts from Google, Adobe, and Teradata. Any opinions, findings, and conclusions or recommendations expressed herein are those of the authors and should not be interpreted as necessarily representing the views, either expressed or implied, of the U.S. Government. The U.S. Government is authorized to reproduce and distribute reprints for government purposes not withstanding any copyright annotation hereon.
\section{Limitations}
The limitations of our method are as follows:
\begin{enumerate}[nosep,leftmargin=*]
\vspace{-2mm}
    \item Our method requires access to a robust knowledge graph to define the concepts and the relationships for distant supervision. 
    \item Our method minimizes the need for but still requires human-annotated data, which is both expensive and time-consuming to create.
    \item The low F1-macro scores of our model and all other leading RE models highlight the need to improve performance on long-tail relation classes in future works. 
\vspace{-2mm}
\end{enumerate}      

\bibliography{anthology,custom}
\bibliographystyle{acl_natbib}
\clearpage
\appendix
\section{Appendix}\label{sec:appendix}

\subsection{Learning order methods: batch- vs epoch-based}\label{app:bb_vs_eb}

\begin{algorithm}
\caption{Batch-based learning order}\label{alg:bblo}
k = 15 epochs\\
\For{$i=0$ \KwTo $k$}{
    \ForEach{batch of training data}{
        predictions $\leftarrow$ model(batch)\\
        $\mathcal{A}_{i}$.insert(correct predictions)\\
        Calculate loss\\
        Back propagate\\
    }
}
\end{algorithm}

\begin{algorithm}
\caption{Epoch-based learning order}\label{alg:eblo}
k = 15 epochs\\
\For{$i=0$ \KwTo $k$}{
    \ForEach{batch of training data}{
        Calculate loss\\
        Back propagate\\
    }
    predictions $\leftarrow$ model(all training data)\\
    $\mathcal{A}_{i}$.insert(correct predictions)\\
}
\end{algorithm}

\subsection{Pre-training Settings} \label{app:pretrain}
We initialize our model with \textit{roberta-base} released by Huggingface\footnote{\url{https://huggingface.co/roberta-base}}. The optimizer is AdamW and we set the learning rate to $3 \times 10^{-5}$, weight decay to $1 \times 10^{-5}$, batch size to 768 and temperature $\tau$ to $5 \times 10^{-2}$. The hyper-parameter $\alpha$ that controls the weights of contrastive learning is $e$ (the base of natural logarithm). We randomly sample 64 negatives for each document. We train our model with 3 NVIDIA Tesla V100 GPUs for 6,000 steps.
\subsection{Downstream Training Settings}
\subsubsection{DocRED} \label{app:docred}
We fine-tune our model on DocRED using the following settings: batch size=32, epochs=200, max sequence length=512, gradient accumulation steps=1, learning rate=4e-5, weight decay=0, adam epsilon=1e-8, max gradient norm=1.0, hidden size=768, and a seed=42. Results are reported on the official DocRED test set as an average of three runs. 

\subsubsection{SemEval and TACRED}\label{app:semeval_tacred}
We fine tune our moodel on SemEval and TACRED using the following settings: batch size=64, max sequence length=100, learning rate=5e-5, adam epsilon=1e-8, weight decay=1e-5, max gradient norm=1.0, warm up steps=500, and hidden size=768. We ran tests on training proportions 0.01/0.1/1.0 using 80/20/8 epochs and a dropout of 0.2/0.1/0.35, respectively.

Results are reported as an average of five runs using the following seed values: 42, 43, 44, 45, and 46.

\begin{figure*}[ht]
    \centering
    \includegraphics[width=.91\textwidth]{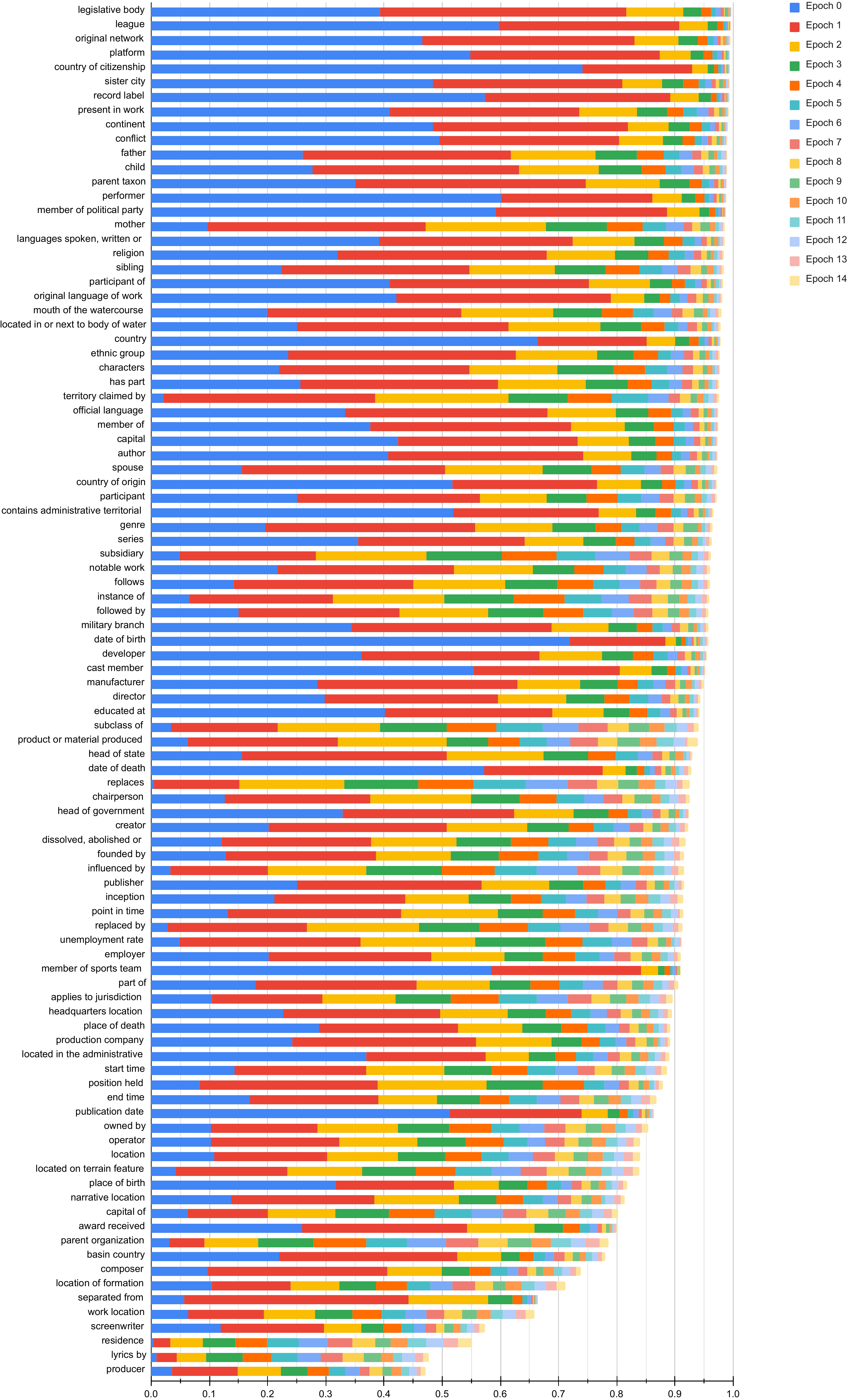}
    \caption{Ratios of instances of learned classes per epoch when recording learning order from distantly supervised DocRED training data. Note, this is before randomized upsampling of underrepresented classes (e.g. \textit{lyrics by} and \textit{producer}). }
    \label{fig:complete_learning_order}
\end{figure*}
\end{document}